\documentclass[letterpaper]{article} % DO NOT CHANGE THIS
\usepackage{aaai23}  % DO NOT CHANGE THIS
\usepackage{times}  % DO NOT CHANGE THIS
\usepackage{helvet}  % DO NOT CHANGE THIS
\usepackage{courier}  % DO NOT CHANGE THIS
\usepackage[hyphens]{url}  % DO NOT CHANGE THIS
\usepackage{graphicx} % DO NOT CHANGE THIS
\usepackage{natbib}  % DO NOT CHANGE THIS AND DO NOT ADD ANY OPTIONS TO IT
\usepackage{caption} % DO NOT CHANGE THIS AND DO NOT ADD ANY OPTIONS TO IT
\usepackage{algorithm}
\usepackage{algorithmic}
\usepackage{tikz}
\usepackage{comment}
\usepackage{amsmath,amssymb} % define this before the line numbering.
\usepackage{color}
\usepackage{booktabs}
\usepackage{wrapfig}
\usepackage{cite}
\usepackage{multirow}
\usepackage{subfigure}

\usepackage{newfloat}
\usepackage{listings}
\DeclareCaptionStyle{ruled}{labelfont=normalfont,labelsep=colon,strut=off} % DO NOT CHANGE THIS
\floatstyle{ruled}
\newfloat{listing}{tb}{lst}{}
\floatname{listing}{Listing}
\def\ie{\emph{i.e}.}

\title{RSPT: Reconstruct Surroundings and Predict Trajectories \\ for Generalizable Active Object Tracking}

\author{
   Fangwei Zhong\equalcontrib \textsuperscript{\rm 1, 2},
   Xiao Bi\equalcontrib \textsuperscript{\rm 3}, 
   Yudi Zhang \textsuperscript{\rm 4}, 
   Wei Zhang \textsuperscript{\rm 4}, 
   Yizhou Wang \textsuperscript{\rm 3, 5}
}
\affiliations{
    \textsuperscript{\rm 1} Sch'l of Intelligence Science and Technology, Peking University\\
    \textsuperscript{\rm 2} Nat'l Key Lab. of GAI, Beijing Institute for General Artificial Intelligence (BIGAI)\\
    \textsuperscript{\rm 3} Center on Frontiers of Computing Studies, Sch'l of Computer Science\\
    Inst. for Artificial Intelligence, Peking University\\
    \textsuperscript{\rm 4} Sch'l of Control Science and Engineering, Shandong University\\
    \textsuperscript{\rm 5} Sch'l of Info. Eng., Zhengzhou University
     
    \{zfw, bixiao, yizhou,wang\}@pku.edu.cn, reed\_zyd@mail.sdu.edu.cn, davidzhang@sdu.edu.cn

}

\newcommand\blfootnote[1]{%
  \begingroup
  \renewcommand\thefootnote{}\footnote{#1}%
  \addtocounter{footnote}{-1}%
  \endgroup
}

\usepackage{bibentry}
\begin{document}

\maketitle

\begin{abstract}
Active Object Tracking (AOT) aims to maintain a specific relation between the tracker and object(s) by autonomously controlling the motion system of a tracker given observations. 
AOT has wide-ranging applications, such as in mobile robots and autonomous driving. However, building a generalizable active tracker that works robustly across different scenarios remains a challenge, especially in unstructured environments with cluttered obstacles and diverse layouts. We argue that constructing a state representation capable of modeling the geometry structure of the surroundings and the dynamics of the target is crucial for achieving this goal. To address this challenge, we present RSPT, a framework that forms a structure-aware motion representation by Reconstructing the Surroundings and Predicting the target Trajectory. Additionally, we enhance the generalization of the policy network by training in an asymmetric dueling mechanism. We evaluate RSPT on various simulated scenarios and show that it outperforms existing methods in unseen environments, particularly those with complex obstacles and layouts. We also demonstrate the successful transfer of RSPT to real-world settings. \blfootnote{Project Website: \url{https://sites.google.com/view/aot-rspt}}.

% Empirical results in virtual environments show that the tracker with RSPT significantly outperforms the existing methods among unseen environments, especially in environments with cluttered obstacles and diverse layouts. We further deploy the RSPT in a real-world scenario, showing good generalization in sim-to-real transfer.
% \keywords{Active Object Tracking, Structure Reconstruction, Trajectory Prediction}
\end{abstract}

% \vspace{-0.1cm}
\section{Introduction}
\label{sec:intro}
% \vspace{-0.1cm}

Active object tracking (AOT) aims to follow a target object by autonomously controlling the motion system of an embodied agent. Specifically, the agent must perceive the movement of the target and its surroundings and subsequently adjust its posture to continuously position the target at the center of its view with an appropriate size. AOT has a vast array of applications, including drones~\cite{ci2023proactive}, mobile robots~\cite{wang2018accurate}, and autonomous driving~\cite{jin2022conquering}.

Although recent years have witnessed remarkable progress in embodied AI~\cite{chrisley2003embodied},
it remains challenging for an agent to actively track a moving object in complex unstructured scenarios~\cite{zhong2019ad}.
% its performance in complex unstructured scenarios is still not ideal. 
\begin{figure}
% \hspace*{0.01\linewidth} \\
\includegraphics[width=\linewidth]{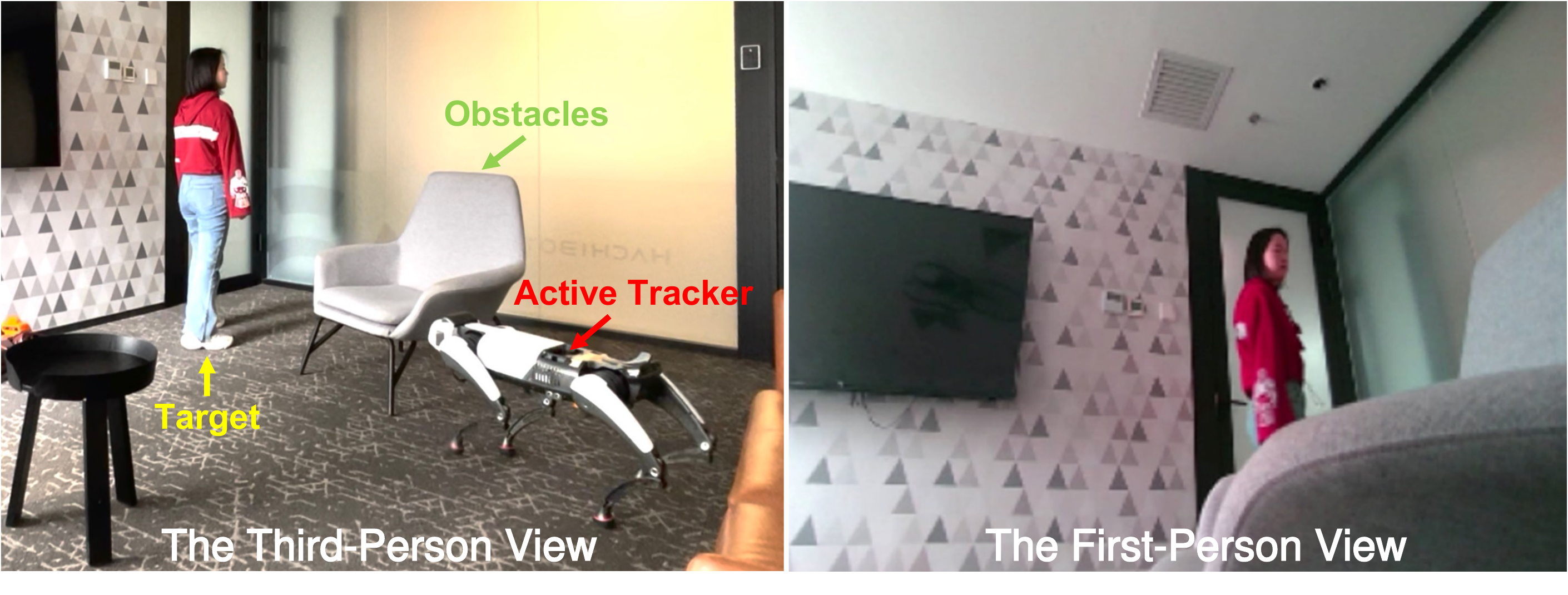}
% \hspace{0.0in}
% \includegraphics[height=0.312\linewidth]{figures/example-sim-small.png}
\caption{This exemplar case highlights the challenges of active object tracking, particularly in environments with cluttered obstacles.
In this scenario, the target (a human) is occluded by obstacles, and the tracker (a robot) must find a way to follow the target without encountering occlusions.
% The left is in real environment, and the right is in the simulation. 
% The target (woman) is occluded by the obstacles and the tracker (robot or man) needs to find a way to follow the target without occlusion.
}
\label{fig:example-1}
\end{figure}
In these environments, obstacles are usually cluttered and arranged in a variety of ways. For example, in a family scene, different families have different room layouts, different furniture placements, and different appearances. The outdoor environment is more complex, \emph{e.g.}, unpredictable obstacles may be encountered at any time. As shown in Figure~\ref{fig:example-1}, the tracker (robot) is required to find a path to follow the target (human) while avoiding occlusion and collision. In these environments, the performance of existing models, especially end-to-end networks~\cite{luo2018end, zhong2018advat}, are dropped significantly.

The inadequate generalization of these models can be attributed to two primary factors: \emph{visual perception} and \emph{physical movement}. The diversity in appearances and layouts of obstacles results in various unseen situations for the tracker. Additionally, target occlusion by obstacles makes it invisible to the tracker, hindering data-driven methods. In terms of physical movement, obstacles may impede the tracker’s path, necessitating route planning to bypass obstacles and re-localize the target.
% We argue that the inadequate generalization of these models can be attributed to two primary factors. With respect to \emph{vision}, the appearances and layouts of obstacles are diverse, resulting in various unseen situations to the tracker. Besides, the target would be occluded by obstacles, making it invisible to the tracker. All these make the current data-driven method hard to deploy. In terms of \emph{physical movement},  obstacles may impede the tracker’s path, necessitating the tracker to plan a safe route to bypass the obstacle and re-localize the target. 

% We contend that the inadequate generalization of these models can be attributed to two primary factors. With respect to vision, the diversity in the appearance and arrangement of obstacles gives rise to a multitude of unprecedented scenarios for the tracker. Moreover, targets may become obscured by obstacles, rendering them undetectable to the tracker. These challenges render current data-driven approaches difficult to implement. In terms of physical movement, obstacles may impede the tracker’s path, necessitating the planning of a safe route to circumvent the obstacle and reacquire the target.

% However, most trackers are lack of geometric common sense for this function.
% when the tracker is blocked by obstacles and affects the movement trajectory, it is difficult for the visual tracker to plan a route to bypass the obstacle to track the target due to the lack of geometric common sense.

To develop a generalizable tracker, it is essential to construct a structure-aware motion representation that meets three requirements: 1) abstraction of the scene’s geometric structure to identify target objects, obstacles, and free space; 2) prediction of the target’s future movement based on geometric structure; and 3) real-time computation of the overall pipeline.

%To build a generalizable tracker, we argue that the key is to construct a structure-aware motion representation, and this representation needs to meet the following requirements: 1) It can provide an abstraction of the geometric structure of the scene, in which the target object, obstacles, and free space can be identified. 2) It can predict the future movement trend of the target based on the geometric structure. 3) The overall pipeline should be computed in real-time.

In this paper, we propose a novel framework for generalizable active object tracking called \textbf{RSPT} (\textbf{R}econstruct \textbf{S}urrounding and \textbf{P}redict \textbf{T}rajectory).
RSPT consists of four modules: Target Localization, Structure Reconstruction, Structure-aware Trajectory Prediction, and a Motion Controller.
The \emph{Target Localization} estimates the 2D location of the target using an off-the-shelf video tracker and transforms it into 3D space with depth images.
The \emph{Structure Reconstruction} module builds a tracker-centric grid map of the environment in real-time using depth images and camera poses.
The \emph{Structure-aware Trajectory Prediction} module integrates historical relative trajectory data with the reconstructed map to predict future target movement.
Finally, the Motion Controller, trained by Reinforcement Learning (RL)~\cite{sutton1998, atari2015}, outputs actions to move the agent step-by-step using the constructed structure-aware motion representation.
We adopt the asymmetric dueling mechanism~\cite{zhong2018advat, zhong2019ad} to improve tracker robustness in complex environments.

% In this paper, we propose a novel framework that can \textbf{R}econstruct the \textbf{S}urrounding of the tracker and \textbf{P}redict the \textbf{T}rajectory of the target to form a structure-aware motion representation for generalizable active object tracking, namely \textbf{RSPT}. There are four modules in RSPT. The first module is the \emph{Target Localization}, which estimates the 2D location of the target via an off-the-shelf video tracker. Then we transform the 2D location into 3D space with the observed depth image. 
% The second module is the \emph{Structure Reconstruction}, which builds a tracker-centric grid map of the environment in real-time with the observed depth images and camera poses. 
% The third module is \emph{Structure-aware Trajectory Prediction} which integrates the historical relative trajectory of the target and the reconstructed map, thereby predicting a variety of walking trends of the target in the future.
% In the end, the \emph{controller}, trained by Reinforcement Learning (RL)~\cite{sutton1998, atari2015} uses the constructed structure-aware motion representation to output actions to move the agent step-by-step.
% Note that we adopt the asymmetric dueling mechanism~\cite{zhong2018advat, zhong2019ad} to further improve the robustness of the tracker in those complicated environments.

We demonstrate that the RSPT-based tracker outperforms previous AOT methods in terms of accumulated reward, episode length, and success rate in both simple and complex environments with cluttered obstacles and diverse layouts. Experimental results show that when obscured or blocked by obstacles, the tracker can predict the target’s movement and plan a path considering surrounding structure to follow the target. Even when the target leaves the field of view, the tracker can retrieve it and continue tracking using memory. Additionally, we deploy the tracker on a real-world robot to demonstrate its strong generalization capabilities both quantitatively and qualitatively.

% We demonstrate that the tracker build on RSPT, compared with the previous AOT methods, can achieve higher accumulated reward, longer episode length, and a higher success rate in both simple environments and complex environments with cluttered obstacles and diverse layouts. From the experimental results, we observe that when the tracker is obscured by obstacles or blocked by obstacles, the tracker can predict the target's towing and plan the path considering the surrounding structure to follow the target. Even if the target has left the field of view, it can also retrieve the target and continue tracking according to the memory.

% We argue that the tracker generated by RSPT, compared with the previous AOT method, can achieve higher accumulated reward, longer episode length, the higher success rate in both simple environments and complex environments with cluttered obstacles and diverse layouts. From the experimental results, we observe that when the tracker is obscured by obstacles or is blocked by obstacles, the tracker can predict the target's towing and plan the path of long-term to follow the target. Even if the target has left the field of view, it can also retrieve the target and continue tracking according to the memory.

The contributions of our work can be summarized as:
\begin{enumerate}
    \item we introduce a structure-aware motion representation that combines the structure of surroundings and dynamic of the moving target in a grid map. This allows for improved tracking route planning in complex environments. %local grid map building in real-time in the AOT task, which makes the tracker have a better perception of the environment.
    \item We present a practical framework that integrates off-the-shelf methods for structure reconstruction, video tracking, trajectory prediction and policy learning to achieve the desired representation.
    % \item We introduce a target trajectory prediction modules to predict a variety of future travel trends of the target, that allows the tracker to better plan tracking routes.
    \item We evaluate trackers in six unseen virtual environments and three real-world robot scenarios, demonstrating strong cross-domain generalization of our RSPT framework. 
    % The results show that our significantly outperforms the state-of-the-art methods.
    % \item We train the environment-target joint awareness via adversarial reinforcement learning based on the above two modules. And we finally produce the generalizable tracker in environments with cluttered obstacles and diverse layouts.
    % \item We deploy RSPT on a real-world quadruped robot and further validate the good sim-to-real transferability with quantitative evaluation.
\end{enumerate}

%-------------------------------------------------------------------------
%\vspace{-0.2cm}
\section{Related Work}
%\vspace{-0.1cm}

%\vspace{-0.1cm}
\textbf{ Active Object Tracking (AOT).} 
%\vspace{-0.1cm}
We can divide the AOT methods into two branches: one-stage methods and two-stage methods.
% The Active object tracking (AOT) methods can be divided into different types naturally, the one-stage modules and the two-stage ones.
Traditional methods perform in the two-stage manner~\cite{kim2005detecting, hong2018virtual}, where the perception module provides a handcrafted state representation, \emph{e.g.}, the 2D bounding box of the target~\cite{ross2008incremental, hu2012single},
then the controller actively adjusts the camera poses accordingly.
Such a solution performs well in most simple cases, however, fails to handle complex cases, \emph{e.g.}, occlusions.
% due to lack of joint adjustment, the perception of target and environments is difficult to balance, thus easily leads to extreme situations.
% delete by bx
% For example, if the tracker focuses on the movement of the target, environmental knowledge will be neglected. If the tracker focuses on reconstructing the complex environment,  a redundant calculation will be introduced and even harm the real-time performance.
% The method of relying only on the results of the passive tracker as input has many limitations. Firstly, multiple modules split need to adjust the parameters for the scene to achieve the optimality of tracking performance, but a set of parameters can only adapt to a specific work scene. Once the robot work scene changes, it is often necessary to re-adjust the parameters. In addition, the traditional method controls the camera movement after obtaining the target's location, it is difficult to consider the complex moving state of the target future. When the camera starts moving, the target often has left the last identified position. In summary, traditional methods are difficult to implement tracking tasks in complex environments. 
% ~\cite{luo2018end, luo2019pami, zhong2019ad, zhong2021distractor}
In recent years, Deep Reinforcement Learning (DRL) is employed by some works~\cite{luo2018end, zhang2018coarse, luo2019pami} to realize AOT tracker in an end-to-end manner.
In the follow-up works ~\cite{zhong2019ad, devo2021enhancing, dionigi2022vat, zhong2021distractor, xi2021anti} demonstrated that multi-agent games can significantly improve tracker's generalization on unseen target, occlusions or distractors.
~\cite{li2020pose} proposed a multi-camera collaboration solution to track the target in complex environments for PTZ camera networks. 
% AD-VAT~\cite{zhong2018advat, zhong2019ad} proposes a novel adversarial reinforcement learning framework to improve the generalization on unseen target or complex environments.
However, it is still difficult to deploy these methods in complex real-world environments, as the sim2real gap.
% Luo, Wenhan's work~\cite{luo2018end}. first proposed end-to-end reinforcement learning methods to solve the active object tracking. This method learned features for input visual information and outputs robot control commands to complete entire processes of active tracking.
% To help the tracker cope with diverse trajectories of the target, AD-VAT~\cite{zhong2018advat} proposes a novel adversarial reinforcement learning framework, where the target learns to escape from the tracker and the tracker learns to catch up with the target. 
% AD-VAT+~\cite{zhong2019ad} uses a two-stage training strategy to improve the robustness of the tracker while tracking with obstacles and occlusions.
% However, they still suffer from collapse while dealing with complex or unseen environments since they only rely on RGB data and hard to obtain the parameters satisfying each sub-function under a one-stage manner. 
% delete by bx
In this work, we aim to build a generalizable tracker that performs well in unseen and complex real-world environments. We focus on state representation and introduce a novel structure-aware motion representation for the tracker. Additionally, we employ an asymmetric dueling mechanism in learning based on AD-VAT.
% In this work, we focus on the building generalizable tracker that can work well on unseen environments, particularly in complex real-world scenarios.
% We investigate the state representation, rather than learning strategy, and construct a novel structure-aware motion representation for tracker.
% Based on AD-VAT, we also employ the asymmetric dueling mechanism in learning.
% Compared with the previous two-stage methods, we follow RL-based methods to optimize our sub-modules jointly under the asymmetric dueling mechanism~\cite{zhong2019ad}.

% mainly focus on the processing of complex target trajectories in an end-to-end manner. 
% Second, these methods relys on RGB-based perception and do not cope with the  our Works mainly focuses on crowded environments and dynamic obstacles and is the first model based on environment-target joint awareness in AOT.

% 端到端泛化能力弱，分步的表征不合适，难以处理复杂场景。
% The above methods 

\textbf{Structure Reconstruction.} 
Map represents the geometry of a scene.
Previous works~\cite{Zhang2014LOAMLO, zhang2015visual, zhong2018detect, campos2021autonomous} have used the point clouds from LiDAR or RGB-D camera as input for map reconstruction.
% Among them, \cite{bonatti2019towards} focuses on the environmental occlusions and collisions.
% delete by bx
% in Unmanned Aerial Vehicle (UAV) autonomous cinematography.
% It uses the LiDAR point clouds to incrementally calculate the local information and represents it as a truncated signed distance field. s
% But the limitation is that LiDAR is too expensive and requires more computation resources, so it is hard to implement in real scenarios. 
However, point clouds are high-dimensional and inefficient for downstream tasks such as policy learning.
Autonomous navigation frameworks~\cite{campos2021autonomous} have been proposed using depth sensors or visual SLAM~\cite{campos2021orb, qin2018vins} in unknown environments, but these methods have limitations such as cumulative errors and high storage requirements.
\cite{usenko2017real} proposed a robot-centric 3D circular buffer to represent the local environment efficiently. 
In AOT, robots rely more on their surroundings than the global environment, making a local robot-centric map more suitable and efficient for tracking.

% Due to the high cost of LiDAR device, other works~\cite{ henry2014rgb, zhong2018detect, campos2021autonomous} use RGB-D as input.
% However, most works use point clouds as the representation of the global map, which is high-dimensional and inefficient for downstream tasks, e.g., policy learning.
% Among them, \cite{campos2021autonomous} provides an autonomous navigation framework, mapping the unknown environment from the disparity measurements obtained from a depth sensor. However, cumulative errors and highly storage requirement are introduced while building a map of the whole environment.
% Visual SLAM~\cite{campos2021orb, qin2018vins} use RGB frames to reconstruct the maps, but those methods are not robust when the scene is lacking texture or dynamic. 
% \cite{usenko2017real} proposes a robot-centric 3D circular buffer to represent the local environment, which is efficient in computation resources and storage space. 
% % This method is more effective both in computation resources and storage space, and even overcomes the dynamic changes of the environment by the real-time updating map.
% % delete by bx
% In AOT, the robots rely on the state of the surrounding environment much more than the global environment. So we argue that the local robot-centric map is more suitable and efficient for the tracker.

\begin{figure*}[!tb]
\centering
\hspace*{0.01\linewidth} \\
\includegraphics[width=0.99\linewidth]{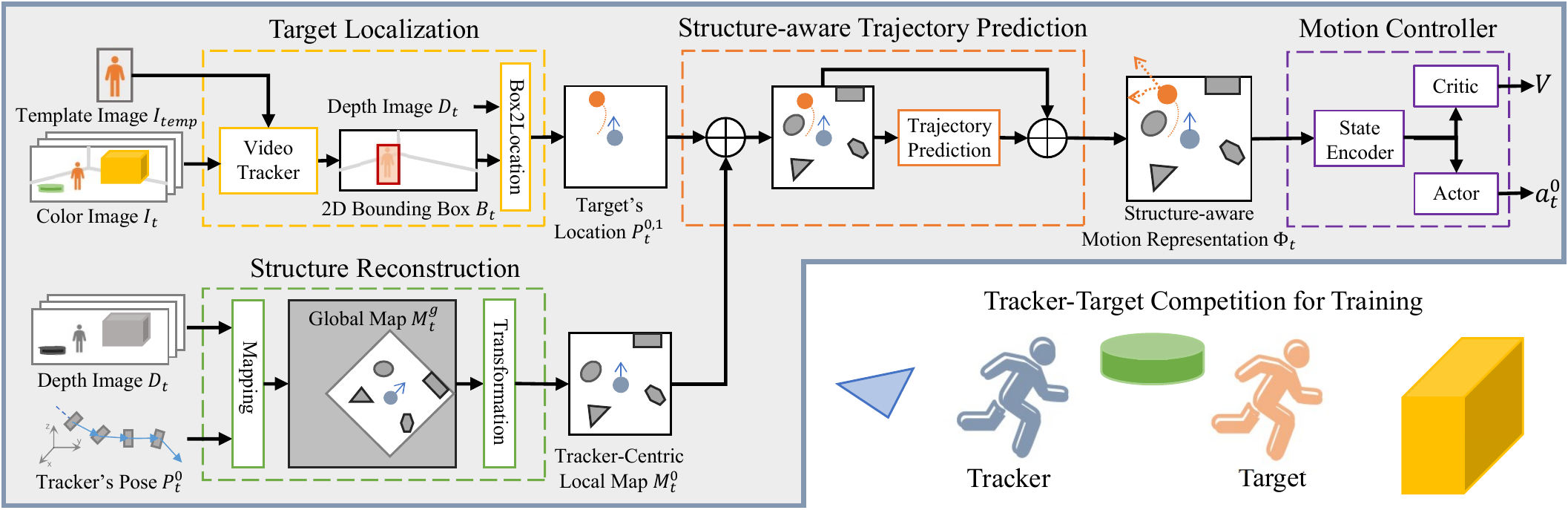}
\caption{An overview of the RSPT framework for active object tracking. It forms a structure-aware motion representation by Reconstructing the Surroundings of the tracker and Predicting the Trajectory of the target. The tracker first localizes the target by a video tracker, and simultaneously constructs a local grid map with the depth image and camera pose, then predicts the future trajectory of the target in the map. 
Based on the representation, the controller learns the tracking policy via Reinforcement Learning (RL) under the asymmetric dueling mechanism~\cite{zhong2019ad}.}
\label{fig:framework}
\end{figure*}

\textbf{Trajectory Prediction.}
Human motion prediction is previously studied in computer vision~\cite{zhao2019multi, tang2019multiple}. 
% Here we mainly pay attention to methods for trajectory prediction by deep learning and combining environmental information. 
% delete by bx
% There are different methods for modeling historical positions. 
% Some works \cite{traj_12, traj_13} put the historical position into the grid map, modeled by Convolutional Neural Networks (CNN)~\cite{nam2016learning}, which converts sequential data into a picture, better representation of spatial position relationship, resulting in increased computation operations. 
\cite{traj_12, traj_13} draw historical position on a grid map to take its sequential and spatial properties into consideration, resulting in increased computation operations.
At present, the main method of historical position still takes it as sequential data, through Recurrent Neural Networks (RNN)~\cite{medsker2001recurrent} to extract the features.
Recently, \cite{traj_b} proposed a method based on Long Short-Term Memory (LSTM)~\cite{hochreiter1997long} to predict the future trajectory of the target over the occupancy grid map. \cite{Alpher10} propose a model based on grid representations to forecast agent trajectories. 
% They use binary 2-D grids, with an agent-centric style to represent the environment and history trajectory information.
They encode the scene and past trajectories using convolutional layers and generate trajectory forecasts using a Convolutional LSTM (ConvLSTM)~\cite{xingjian2015convolutional}. 
% delete by bx
In AOT, the tracker is required to predict the target’s motion from a first-person view. To account for cluttered obstacles, we integrate scene structure and historical trajectory of the target to forecast the target’s trajectory distribution.

% \vspace{-0.1cm}
\section{Method}
% \vspace{-0.1cm}

The RSPT framework aims to create a structure-aware motion representation, denoted by $\Phi_t$, using the RGB image $I_t$, the depth image $D_t$, and the camera pose of the tracker, $P^0_t$. 
The framework consists of four main components, namely, \emph{target localization}, \emph{structure reconstruction}, \emph{structure-aware trajectory prediction}, \emph{motion controller}, as shown in Figure~\ref{fig:framework}. By using this state representation, a generalizable tracker can be trained through reinforcement learning (RL) in an asymmetric dueling mechanism, as proposed in AD-VAT ~\cite{zhong2018advat}.

% \vspace{-0.1cm}
\subsection{Target Localization}
% \vspace{-0.1cm}

The most basic function of the tracker is to identify and localize the target.
In this regard, we employ an off-the-shelf video tracker, denoted by $VT$, to estimate the 2D location of the target object in the image space. 
Specifically, we first crop the first RGB image $I_0$ according to the initial target bounding box $B_0$, which serves as a template denoted by $I_{temp}$.
Then, we use $VT$ to detect the bounding box $B_t$ of the target object in the subsequent RGB image $I_t$. To improve cross-domain generalization, we adopt DiMP~\cite{bhat2019learning}, which can online tune the network parameters to better fit the target appearance during testing.
The overall formula can be expressed as follows:
% \vspace{-0.6cm}
% \begin{align}
% I_{temp} = crop(i_0, b_0).
% \end{align}
% \vspace{-0.6cm}

% \vspace{-0.6cm}
\begin{align}
B_t = VT(I_t, I_{temp}).
\end{align}
% \vspace{-0.6cm}
% Specifically, at each time step, we get color images from the camera and the template of the target from the initialization. 
% delete by bixiao
% Then we use the DiMP~\cite{bhat2019learning} as the video tracker, to provide the 2D bounding box of the target. 
% Of course, other video trackers, \emph{e.g.}, SINT~\cite{tao2016siamese}, ATOM~\cite{danelljan2019atom}, SiameseRPN~\cite{zhu2018distractor}, SiamRPN++~\cite{li2019siamrpn++} and SiamFC~\cite{bertinetto2016fully}, can also be employed. 
% For a better cross-domain generalization, we adopt DiMP~\cite{bhat2019learning} which can online tune the network parameters to fit the target appearance while testing.

To model the historical trajectory of the target, it is necessary to transform the bounding box $B_t$ in image space into the target's location $P^{0,1}_t$ in the camera coordinate system. Specifically, we obtain the central location $(u, v)$ of the bounding box $B_t$ and the corresponding depth value $d$ in the depth image $D_t$. Next, we use the camera intrinsic parameters $K$ to convert the $(u, v, d)$ triplet into the target's 3D relative position $P^{0,1}_t$. The formula for this transformation is as follows:

% \vspace{-0.6cm}
\begin{align}
P^{0,1}_t = CT(K, D_t, B_t).
\end{align}
% \vspace{-0.6cm}

% We get the target's bounding box $B_t$ central location $UV$ and the corresponding point depth $D$. Then, using the camera intrinsic $K$, convert the $UVD$ triplet to the 3D relative position.
The coordinate transforms $CT$ is used to obtain the relative pose of the target with respect to the tracker. Here $K$, $D_t$ and $B_t$ represent the camera intrinsic parameters, the depth image, and the 2D bounding box of the target, respectively.
The coordinates $(x, y)$ represent the position $P^{0,1}_t$ of the target relative to the tracker, where $x$ and $y$ correspond to the lateral and longitudinal directions, respectively. At each time step, we set the tracker's coordinate as $(0, 0)$, and the target's coordinate at time step $t$ is denoted by $(x_t, y_t)$.

To help the tracker in inferring the current location of the target, we define the visibility state of the target $Vis_t^1 = [P^{0,1}_t, P^{1}_{last}, C_{inv}]$. Specifically, $P^{0,1}t$ represents the current position of the target relative to the tracker, $P^{1}_{last}$ represents the last observed position of the target, and $C_{inv}$ denotes the number of consecutive time steps during which the target has not been observed by the tracker. In instances where the target is not visible, the video tracker outputs a "not found" flag and sets the relative position $P^{0,1}_t$ to all zeros. Notably, all position values are relative to the tracker.
%  \vspace{-0.1cm}
\subsection{Structure Reconstruction}
% \vspace{-0.1cm}
The aim of the structure reconstruction module $SR$ is to reconstruct the environment in real-time to represent the recent state of the tracker's surroundings. Visual SLAM systems~\cite{campos2021orb} can perform this task, but their maps are not suitable for downstream tasks in real time. This is because processing 3D point clouds requires heavy computation, and sparse keypoint clouds provided by most SLAM systems lack necessary information for decision-making, such as the absence of points in a blank wall.

Motivated by this, we utilize a grid map \cite{8392399} that is based on the depth image $D_t$ and the estimated pose $P^0_t$ at each time step. 
% The localization system, equipped in most robots, can output the pose. In the virtual environment, we add random noise to the grounded pose to simulate the real signal.
To reduce computation costs, we recover point clouds within a certain height from the depth image and camera intrinsic parameters $K$. This approach may lead to a loss of map precision, especially in scenarios with large variations in ground height and variable obstacle sizes. Nevertheless, we argue that this trade-off between accuracy and computational efficiency is acceptable. Empirical results demonstrate that such maps are generally suitable for downstream decision-making tasks.
We then update the occupancy probability of each grid cell based on the density of the corresponding point cloud, allowing the grid map to reflect the probability of being occupied by obstacles, including dynamic ones like the target.
Additionally, the 2D grid map is more amenable to real-time computation by convolutional neural networks, which is necessary for downstream tasks such as trajectory prediction and motion control. 
We combine obstacle information seen at different times with the tracker pose $P^0_t$, and finally obtain the tracker-centric local map $M_t^0$ by transforming the grid-based map to the current pose coordinate. 
% When using raw RGB-D, the surrounding structure is incomplete.
% Finally, the tracker-centric local map $M_t^0$ can be acquired by transforming the local grid-based map to the current pose coordinate. 
Note that the pose can be obtained from a localization system equipped in most robots or simulated by adding noise to a grounded pose in virtual environments.
% Note that the localization system, equipped in most robots, can output the pose. In virtual environments, we add noise to simulate real-world localization signals.
The overall formula is as follows:
% \vspace{-0.2cm}
\begin{align}
M_t^0 = SR(D_t, K, P^0_t).
\end{align}
% \vspace{-0.6cm}
 
The local grid map is partitioned into fixed-size rectangular grids of equal size. To ensure a valid distance range of the RGB-D sensors and avoid accumulating errors, we limit the coordinate of the surrounding structure to within -10 to 10 meters. In our method, we employ a grid size of $(40 \times 40)$, with each grid spanning $50cm$ in both length and width. This grid-based representation can be efficiently processed by a Neural Network for real-time decision making.

% Such local grid map divides the region around the tracker into fixed-size rectangular grids, and each grid is the same size. We consider the coordinate of the surrounding structure in the range of -10 to 10 meters, considering the valid distance range of the RGB-D sensors and avoidance of cumulative errors. In our system, we use $(40 \times 40)$ grids, where the length of each grid spans $50cm$ and the width of each grid also spans $50cm$. Such grid-wise map can be easily encoded by the Convolutional Neural Network (CNN) in real-time for the decision making. 

% \vspace{-0.2cm}
\subsection{Structure-aware Trajectory Prediction}
% \vspace{-0.1cm}
To predict the trajectory of the target on the map, we employ a trajectory predictor $TP$ to estimate the target trajectory $\tau^1_{t+1:t+F}$. At each time step, we use the most recent $H$ samples of the target's coordinates, $\tau^1_{t-H+1:t} = [(x_{t-H+1}, y_{t-H+1}), ..., (x_t, y_t)]$, to forecast the $F$ subsequent coordinates, $\tau^1_{t+1:t+F} = [(x_{t+1}, y_{t+1}), ..., (x_{t+F}, y_{t+F})]$. We formulate this module as follows:

\begin{align}
\tau^1_{t+1:t+F} = TP(E_{map}(M_t^0), E_{traj}(\tau^1_{t-H+1:t})).
\end{align}
% Each coordinate is acquired every $T$ seconds and the current target's position is available. 

% We obtain the coordinate information of the target by processing the measurements from the RGB-D camera sensor. At each time, we get images from the camera, then get the pixel coordinates of the target by the passive tracker, and then convert them to the coordinate of the camera coordinate system through the internal parameters of the camera.

We employ Convolutional Neural Networks (CNN) as a feature extractor $E_{map}$ to capture the spatial surrounding structure. For the sequential information of the target's historical location $\tau^1_{t-H+1:t}$, we use Long Short-Term Memory (LSTM) as a feature extractor $E_{traj}$. In order to predict the target's trajectory, we compare the performance of simple regression and Mixture-Density Recurrent Network~\cite{8500419, Makansi_2019_CVPR} (MDN-RNN). Our experiments demonstrate that regression can only model the average distribution of the trajectories, whereas MDN-RNN can capture the distribution of the trajectories and generate multiple possible future trajectories. Specifically, for each future time step, MDN-RNN outputs the core weights, the mean and variance of the relative coordinate $x$ and $y$ in each GMM core, and concatenates them to form the target trajectory $\tau_t^1$. The final trajectory distribution is then computed accordingly. 
% target trajectory $Traj_t^1$ is finally combined with the weights of $K$ GMM cores, and mean and variance values of the $x$ and $y$ coordinates for each core in $F$ sampled coordinates $P^1_{fut}$. The calculation of trajectory predictor $g_{o}$ can be represented as
% \vspace{-0.6cm}
Our PT module has two key advantages. Firstly, it conditions the trajectory prediction on the spatial distribution and size of obstacles in the environment, leading to more accurate trajectory forecasts in cluttered settings. Secondly, it estimates the likelihood of multiple potential paths, allowing the tracker to react more flexibly and rapidly to changes in the target's behavior, rather than assuming a regular movement pattern. We validate the effectiveness of our PT module by comparing it against a Kalman Filter, as detailed in Section \ref{subsec:exp-baseline}.

Then, we construct a structure-aware motion representation that comprises the visibility state of the target, the tracker-centric local map, and the predicted trajectory, denoted as $\Phi_t = [Vis_t^1, E'_{map}(M_t^0), E'_{traj}(\tau_t^1)]$. Notably, the encoder $E'_{map}$ and $E'_{traj}$ employ the identical network architecture as $E_{map}$ and $E_{traj}$, respectively, but possess independent network parameters for learning.
% For feature extraction of the surrounding structure $M_t^0$, we use the classic visual feature extractor $E_c$ with Convolutional Neural Networks (CNN). We use feature extractor $E_r$ with Long Short-Term Memory (LSTM) for target trajectory $Traj_t^1$. 
% The formula of structure-aware motion representation is 
% \vspace{-0.2cm}
% \begin{align}
% \Phi_t = [Vis_t^1, E_c(M_t^0), E_r(Traj_t^1)].
% \end{align}

% Our PT module based on MDN-RNN offers two benefits. First, it predicts the trajectory conditioned on the size and location of the obstacles, resulting in a more accurate forecast in crowded surroundings.
% such as the future trajectory of the target will not pass through obstacles.
% Second, it forecasts the likelihood of different paths, allowing the tracker to respond flexibly and swiftly rather than assuming that the target constantly moves in a regular curve. We test this by replacing our PT module with a Kalman Filter which will be mentioned in \ref{subsec:exp-baseline}.

% \vspace{-0.1cm}
\subsection{Motion Controller}
% \vspace{-0.1cm}

% We get a real-time grid map from the structure reconstruction module and future trajectory distribution from the trajectory prediction module. 
We propose a neural network-based motion controller, denoted by $a_t = MC(\Phi_t)$, which takes the structure-aware motion representation $\Phi_t$ as input and outputs the action for the tracker at time step $t$. The motion controller is trained through reinforcement learning~\cite{luo2019pami}, where the objective is to maximize the expected cumulative reward over a finite horizon.

% \vspace{-0.6cm}
To enhance tracker robustness, we employ the asymmetric dueling mechanism~\cite{zhong2019ad} to automatically generate diverse target trajectories while training.
Specifically, the target is an end-to-end network, and its input is the grounded location of all agents obtained in the simulation. This allows the target to identify weaknesses in the tracker's movements, thereby learning from successful escape paths.
The reward structure is based on AD-VAT+~\cite{zhong2019ad}. Specifically, the reward for the tracker is defined as $r_0 = 1 - \Delta\rho - \Delta\theta$, where $\Delta\rho$ and $\Delta\theta$ denote the errors in relative distance and angle, respectively. As the target approaches the expected position, the tracker's reward increases. The target's reward is the negative of the tracker's reward written as $r_1 = -r_0$, incentivizing the target to evade the tracker. By training with an adversarial target, the robustness of the tracking policy is improved concurrently.

\section{Experiments}
% \vspace{-0.1cm}

\label{sec:exp}
In this section, we conduct experiments to address the following questions: 1) How does the performance of the proposed framework compare with that of existing methods? 2) To what extent does each module contribute to the overall improvement of the framework? 3) What is the primary bottleneck for active object tracking? 4) Can the proposed framework effectively handle noisy observations? 5) Can the framework be successfully deployed in real-world robotic applications? These experiments provide insights into the effectiveness and limitations of our approach, as well as its potential for practical applications.

% In this section, we conduct experiments to answer the following questions: 1) How does the performance of the proposed framework compare with that of existing methods? 2) To what extent does each module contribute to the overall improvement of the framework? 3) What is the primary bottleneck for active object tracking? 4) Can the proposed framework effectively handle noisy observations? 5) Can the framework be successfully deployed in real-world robotic applications? 
% The experiments entail a rigorous analysis of the framework's accuracy, efficiency, and robustness across various evaluation metrics and datasets, using established benchmarks in the field of active object tracking. The results are discussed in detail, providing insights into the strengths and limitations of the proposed framework.

% In this section, we conduct experiments to answer the following questions:
% 1) How is the framework compared with baselines? % (Comparison with baselines)
% 2) Does each module contribute to the improvement of the framework?
% % (Ablation study) 
% 3) What is really the bottleneck for active object tracking? 
% % (Running with grounded policy)
% 4) Can the framework handle the noisy observation? % (Running with Noisy Input)
% 5) Can the framework deploy on the real-world robot? % (Real-world Deployment)

The details of the experiments are introduced as follows. 

\begin{figure*}[t]
\centering
\hspace*{0.01\linewidth} \\
\includegraphics[width=0.99\linewidth]{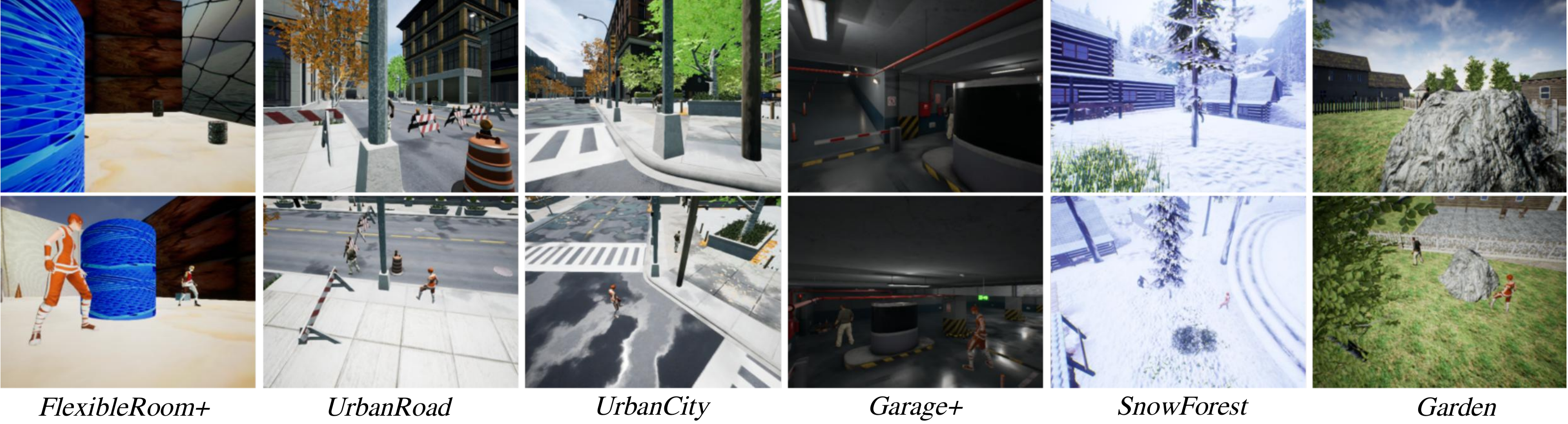}
% \vspace{-0.2cm}
\caption{Examples of the environments for training (leftmost) and testing (others). The top row features first-person views of the trackers, while the bottom row displays third-person views}
\label{fig:environments}
% \vspace{-0.5cm}
\end{figure*}

% \vspace{-0.5cm}
\subsection{Environments}
% \vspace{-0.1cm}

We employ six high-fidelity environments from AD-VAT+~\cite{zhong2019ad} to evaluate our approach, as shown in Figure~\ref{fig:environments}. These environments contains diverse obstacles with varying shapes and layouts among these environments. 
The \emph{FlexibleRoom+} environment is used for training, where we randomize the layouts, shapes, and sizes of obstacles in each episode. The remaining six realistic environments (\emph{UrbanCity}, \emph{UrbanRoad}, \emph{Garden}, \emph{Garage}, \emph{Garage+}, and \emph{SnowForest}) are utilized to evaluate the generalization of the trackers in unseen environments. 
%补上action space，observation space
The action space is akin to \cite{zhong2018advat}, with seven discrete actions: \emph{Move-forward}, \emph{Turn-left}, \emph{Turn-left-and-move-forward}, \emph{Turn-right}, \emph{Turn-right-and-move-forward}, \emph{Stop}, and \emph{Move-backward}.
% Forward (Move-forward), Left (Turn-left, Turn-left-and-move-forward), Right (Turn-right,  Turn-right-and-move-forward), Stop (No-op), and Backward (Move-backward). Please refer to the supplementary material for the details of above environments.

% The advantages for training and evaluating in these environments are three folds. Firstly, all of them are constructed based on Unreal Engine 4, which can simulate the physical properties of the real scene properly. Secondly, UnrealCV~\cite{qiu2017unrealcv} provides flexible interfaces for interaction, making us easily access the environment to get the states for observation and reward calculation or control the agents. Thirdly, it can help us avoid the high cost of trial-and-error learning in the real scene. 

% In the following, we described each environment in detail and showed how they place challenges to AOT.

\begin{table*}[t]
\centering

% \tiny
% \vspace{-0.2cm}
\resizebox{\linewidth}{!}{
\begin{tabular}{c|cccccc|c}
\hline
Methods
& \emph{UrbanCity} & \emph{UrbanRoad}    
& \emph{Garage}    & \emph{Garage+}      
& \emph{Garden}    & \emph{SnowForest}   
& Mean      % & Std   
\\ \hline
VT + Rule
& 263/428/0.72 & 250/452/0.75 
& 261/418/0.74 & 223/385/0.67 
& 208/381/0.61 & 119/301/0.43 
& 221/394/0.65 % & 54/53/0.12 
\\ 

VT + Planner
& 215/476/0.84 & 200/ 464/0.84
& 145/445/0.70 &162/421/0.66 
&173/\textbf{450}/0.74 & 72/ 343/0.46 
&161/433/0.71 % &  46/43/0.13  
\\ 

VT + RL
% & 187/359/0.52 & 153/267/0.34 
% & 95/282/0.40  & 64/199/0.31  
% & 173/299/0.31 & 134/222/0.40 
% & 134/271/0.38 & 42/52/\textbf{0.07}  \\ 

& 363/500/1.00 & \textbf{355}/490/0.94 
& 284/427/0.70 & 239/393/0.56  
& 234/394/0.56 & 152/295/0.38 
& 271/416/0.69 %& 73/68/0.22  
\\

Seg + VT + RL
& 303/437/0.83 & 311/469/0.91 
& 283/425/0.65 & 243/416/0.63 
& 205/403/0.67 & 232/364/0.59 
& 263/419/0.71 % & 42/35/0.13 
\\

\hline
AD-VAT 
& 335/484/0.88 & 246/429/0.60 
& 86/302/0.20  & 39/273/0.10  
& 112/297/0.16 & 169/364/0.44 
& 164/358/0.39 %& 100/76/0.27 
\\

AD-VAT+  & \textbf{389}/497/0.94 & 326/471/0.80  & 267/439/0.60 & 166/366/0.40 & 108/277/0.16 & 182/365/0.44 & 240/402/0.55 %& 97/74/0.25  
\\  

TS  &  341/496/0.94 & 308/480/0.84  & 265/472/\textbf{0.89} & 237/\textbf{462}/\textbf{0.78}  & 55/209/0.02 & 234/\textbf{424}/0.63 & 240/424/0.68 % & 100/99/0.34  
\\  

\hline

RSPT (Ours)  & 341/\textbf{500}/\textbf{1.00} & 346/\textbf{500}/\textbf{1.00} & \textbf{314}/\textbf{480}/0.80 & \textbf{285}/445/0.73 & \textbf{283}/444/\textbf{0.72} & \textbf{248}/410/\textbf{0.80} & \textbf{302}/\textbf{463}/\textbf{0.84} % & \textbf{34}/\textbf{32}/\textbf{0.11}  
\\ \hline
\end{tabular}}
\caption{The quantitative results compared with the baselines, where the best results are shown in bold. Note that the top four methods are two-stage methods. VT and Seg represent two kinds of perception modules, \ie, VT = Video Tracker, Seg = Semantic Segmentation. Rule, Plan, and RL represent three kinds of controllers. TS indicates an end-to-end training tracker with the teacher-student learning strategy introduced in ~\cite{zhong2021distractor}. 
Note that all the end-to-end methods (AD-VAT, AD-VAT+, and TS) use the RGB-D image as the input.
The three numbers in each cell represent Accumulated Reward (AR), Episode Length (EL), and Success Rate (SR) respectively.
}
\label{table-baseline}
% \vspace{-0.3cm}
\end{table*}

\begin{table*}[t]
\centering
% \vspace{-0.3cm}
\resizebox{\linewidth}{!}{
% \tiny
\begin{tabular}{c|cccccc|c}
\hline
Methods
& \emph{UrbanCity}    & \emph{UrbanRoad}    
& \emph{Garage}       & \emph{Garage+}      
& \emph{Garden}       & \emph{SnowForest}   
& Mean            % & Std    
\\ 
\hline
VisT      
& 239/401/0.66 & 168/308/0.33 
& 166/327/0.48 & 100/215/0.40
& 154/315/0.34 & 128/278/0.53 
& 159/307/0.45 %& 42/55/0.11 
\\

VisT + RS 
& 313/\textbf{500}/\textbf{1.00} & 305/\textbf{500}/\textbf{1.00} 
& 284/477/0.76 & 249/431/0.69 
& 247/438/0.70 & 171/365/0.60 
& 261/451/0.79 %& 47/47/0.15 
\\

VisT + PT 
& 335/\textbf{500}/\textbf{1.00} & 316/451/0.80 
& 311/472/0.78 & 244/390/0.63 
& 277/417/0.68 & 184/380/0.61 
& 277/435/0.75 %& 51/43/0.13 
\\ 

\hline

RSPT (MDN-RNN)
& \textbf{341}/\textbf{500}/\textbf{1.00} & \textbf{346}/\textbf{500}/\textbf{1.00} & \textbf{314}/\textbf{480}/\textbf{0.80} & \textbf{285}/\textbf{445}/\textbf{0.73} & \textbf{283}/\textbf{444}/\textbf{0.72} & \textbf{248}/\textbf{410}/\textbf{0.80} & \textbf{302}/\textbf{463}/\textbf{0.84} %& \textbf{34}/\textbf{32}/\textbf{0.11} 
\\

RSPT (KF)
& 329/\textbf{500}/\textbf{1.00} & 318/\textbf{500}/\textbf{1.00}
& 284/459/0.70 & 262/439/0.71 
& 251/440/0.68 & 219/387/0.65 
& 277/454/0.79 %& 42/43/0.16  
\\  
\hline
\end{tabular}}
\caption{The quantitative results compared with the ablations, where the best results are highlighted in bold.
VisT denotes the visibility state of the target, RS represents Reconstruct Structure, PT indicates Predict Trajectory, and KF is a Kalman Filter that offers similar functionality to our PT module. Each cell displays three numbers, which represent the Accumulated Reward (AR), Episode Length (EL), and Success Rate (SR), respectively.}
\label{table-ablation}

\end{table*}

\begin{table*}[t]
\centering

% \vspace{-0.1cm}
\resizebox{\linewidth}{!}{
% \tiny
\begin{tabular}{c|cccccc|c}
\hline
\textbf{} & \emph{UrbanCity}    & \emph{UrbanRoad}    & \emph{Garage}       & \emph{Garage+}      & \emph{Garden}       & \emph{SnowForest}   & Mean   %  & Std        
\\ \hline
Object Mask & 302/494/0.98 & 353/499/0.94 & 177/378/0.52 & 232/420/0.64 & 162/389/0.38 & 37/264/0.01  & 210/407/0.57% & 102/79/0.33 
\\
Target Mask  & \textbf{407}/\textbf{500}/\textbf{1.00} & \textbf{392}/494/0.96 & \textbf{359}/474/0.84 & 218/444/0.66 & 377/497/0.98 & 305/461/0.83 & 343/478/0.87 %& 64/20/0.11  
\\ 
RSPT with GT   & 375/\textbf{500}/\textbf{1.00} & 356/\textbf{500}/\textbf{1.00} & 347/\textbf{493}/\textbf{0.90} & \textbf{350}/\textbf{479}/\textbf{0.84} & \textbf{385}/\textbf{500}/\textbf{1.00} & \textbf{384}/\textbf{500}/\textbf{1.00} & \textbf{366}/\textbf{492}/\textbf{0.93} 
%& \textbf{15}/\textbf{8}/\textbf{0.06}  
\\ \hline
% RSPT+ND   & 350/\textbf{500}/\textbf{1.00} & 331/473/0.82 & 281/445/0.73 & 264/411/0.68 & 259/417/0.69 & 204/375/0.62 & 281/436/0.75 & 48/41/0.12 \\
% RSPT+NP   & 339/\textbf{500}/\textbf{1.00} & 347/\textbf{500}/\textbf{1.00} & 305/477/0.77 & 262/429/0.64 & 266/436/0.70 & 229/404/0.78 & 291/457/0.81 & 42/36/0.13 \\ \hline
RSPT (Ours) & 341/\textbf{500}/\textbf{1.00} & 346/\textbf{500}/\textbf{1.00} & 314/480/0.80 & 285/445/0.73 & 283/444/0.72 & 248/410/0.80 & 302/463/0.84 %& 34/32/0.11 
\\ \hline
\end{tabular}}
\caption{The quantitative results with different grounded state. The top two methods take Object Mask, and Target Mask as input, separately, while the RSPT with GT method replace the estimated pose in RSPT with the grounded pose. Note that the RSPT with GT methods provides an upper bound of our RSPT methods.
The three numbers in each cell represent Accumulated Reward (AR),
Episode Length (EL), and Success Rate (SR) respectively.}
\label{table-gt}
\end{table*}

\begin{table*}[t]
\centering
% \vspace{-0.2cm}
\resizebox{\linewidth}{!}{
% \tiny
\begin{tabular}{c|cccccc|cc}
\hline
\textbf{Noisy Input} & \emph{UrbanCity}    & \emph{UrbanRoad}    & \emph{Garage}       & \emph{Garage+}      & \emph{Garden}       & \emph{SnowForest}   & Mean     % & Std        
\\ \hline
% Object Mask+Depth & 302/494/0.98 & 353/499/0.94 & 177/378/0.52 & 232/420/0.64 & 162/389/0.38 & 37/264/0.01  & 210/407/0.57 & 102/79/0.33 \\
% Mask+Depth  & \textbf{407}/\textbf{500}/\textbf{1.00} & 392/494/0.96 & 359/474/0.84 & 218/444/0.66 & 377/497/0.98 & 305/461/0.83 & 343/478/0.87 & 64/20/0.11  \\ 
% RSPT+GT   & 375/\textbf{500}/\textbf{1.00} & 356/482/0.87 & 347/493/0.90 & 350/479/0.84 & 385/\textbf{500}/\textbf{1.00} & 384/\textbf{500}/\textbf{1.00} & 366/492/0.93 & 15/8/\textbf{0.06}  \\ \hline
Clean Input & 341/500/1.00 & 346/500/1.00 & {314}/{480}/0.80 & 285/445/0.73 & 283/444/0.72 & 248/410/0.80 & 302/463/0.84 %& 34/32/0.11 
\\
Depth   & 350/{500}/{1.00} & 331/473/0.82 & 281/445/0.73 & 264/411/0.68 & 259/417/0.69 & 204/375/0.62 & 281/436/0.75 % & 48/41/0.12 
\\
Pose   & 339/{500}/{1.00} & 347/{500}/{1.00} & 305/477/0.77 & 262/429/0.64 & 266/436/0.70 & 229/404/0.78 & 291/457/0.81 % & 42/36/0.13 
\\
Depth + Pose &
336/500/{1.00}&
328/466/0.81&
244/407/0.62&
265/423/0.66&
254/402/0.64&
198/368/0.57&
270/427/0.71
% &48/43/0.14
\\
 \hline
\end{tabular}}
\caption{Evaluating RSPT with different noisy inputs. The four methods take grounded information, noise depth + grounded pose, grounded depth + noise pose, and noise depth + noise pose as input, separately. The three numbers in each cell represent Accumulated Reward (AR),
Episode Length (EL), and Success Rate (SR) respectively.}
\label{table-noise}
\end{table*}

% \vspace{-0.1cm}
\subsection{Evaluation Metric}
% \vspace{-0.1cm}
In our experiment, we set the maximum length of each episode to $500$ time steps.
Besides, a visible area is defined as a fan shape area in front of the tracker with a radius of $750 cm$ and a range of $90$ degrees. The target is considered lost if it moves outside of this visible area, and the episode terminates if the target remains lost for more than $5$ seconds. 
To evaluate the tracking performance, each method runs over $50$ episodes in each environment, and we report average results for three metrics: Accumulated Reward, Episode Length, and Success Rate.
1) \textbf{Accumulated Reward (AR).} This metric, consistent with other RL tasks, measures the cumulative reward across an entire episode, and reflects both accuracy and robustness of the tracker. 
2) \textbf{Episode Length (EL).} We calculate the average episode length based on the defined episode termination condition, which reflects the long-term tracking performance.
% It is worthy to notice that the maximum value of EL is 500.
3) \textbf{Success Rate (SR).} An episode is considered successful if its length reaches the maximum of $500$ steps. We calculate the rate of successful episodes across all episodes, which serves as a measure of model robustness. Higher values indicate greater robustness.

% We report the average value for each evaluation metric over $50$ episodes of each method.

% \vspace{-0.1cm}
\subsection{Comparison with Baselines}
\label{subsec:exp-baseline}
% \vspace{-0.1cm}

% In the following, we compare our model with three well-designed baselines and the results show the excellent performance of our methods.
In the following, we compare our model with four two-stage methods and two end-to-end methods in six realistic unseen environments. 
Results are shown in Table~\ref{table-baseline}.
 % According to the results, our model achieved much better performance compared with the baseline method nearly among all the environments.
% Different from the original version, we extend tracker models by extending the observation from RGB input to RGB-D input for a fair comparison, which will be noted as $*$.
% and the results show the excellent performance of our methods.

% \vspace{-0.1cm}
\textbf{Comparison with Two-stage Methods.}
% We conduct multi two-stage methods to serve as baseline methods.
In the two-stage methods, there are a visual perception module and a controller. For perception, we use an off-the-shelf video tracker (VT) or a semantic segmentation model (Seg) to encode the raw-pixel observation into a state representation, \emph{e.g.}, the bounding box of target or the pixel-level segmentation mask.  % In RSPT, we use the same video tracker for target localization.
Specifically, we employ pretrained DiMP~\cite{bhat2019learning} to track objects in videos and UNet + FCN from MMSegmentation\footnote{\url{https://github.com/open-mmlab/mmsegmentation}} for semantic segmentation.
% \textbf{Seg} means that an off-the-shelf semantic segmentation method is used for scene perception. The semantic segmentation module we use is from MMSegmentation\footnote{\url{https://github.com/open-mmlab/mmsegmentation}}. 
For control, We build three kinds of control strategy, namely Rule, Planner, and RL.
To be specific, the \textbf{Rule}-based controller is a PID-like controller, which outputs action based on the errors between the target location and the expectation. The \textbf{Planner} will reconstruct the surrounding map and uses the path planning algorithms, \emph{e.g.}, A*~\cite{hart1968formal} to navigate the tracker to the expected distance next to the target. The \textbf{RL}-based method is trained via deep reinforcement learning to handle high-dimensional state representation.

 First, we can see that RL-based controller achieved the best performance among the three VT-based method (\emph{VT + Rule}, \emph{VT+ Planner}, \emph{VT + RL}), showing the good tracking performance of the RL-based method. The \emph{VT + Planner} adds path planning to achieve obstacle avoidance. However, it always has a delay, leading to the worst performance. Intuitively, when the tracker plans the path and starts walking according to optimal location in the current state, the target has left its previous position. This makes it difficult to maintain a specific distance and angle with the target. We also notice that \emph{VT + RL} and \emph{Seg + VT + RL} achieve a comparable results, showing that additional semantic information about the scene is useless to the tracker.
 % methods perform poor in obstacle environments.
 % Comparing the results, we can see that VT + Rule, VT + RL, and Seg + RL act only relies on the target's location in 2D image space at the current time without considering the surrounding structure and target trajectory trend. 

% \textbf{VT + Rule} adds control rules after Video Tracker. Specifically, the tracker will act according to the target's offset from the image center and the relative size maintaining size and viewing angle to the target as constant. 
% % For example, the target is on the right side of the picture, then the tracker will go right. If the target's bounding box size is reduced, the tracker will go forward. 
% \textbf{VT + Plan} adds path planning after Video Tracker. First, it reconstructs the surrounding map in real-time. Then, it uses Video Tracker and the camera intrinsic to get the target's location. Finally, it uses the path planning method to navigate to the expected distance next to the target. Here the path planning method we use is A*～\cite{hart1968formal}.
% \textbf{VT + RL} gets the target's location like VT + Plan, and further constructs a Reinforcement Learning (RL) based controller to output the actions.

% \textbf{Seg + RL} is Segmentation with Depth baseline which is based on \cite{hong2018virtual} and takes segmentation mask from MMSegmentation\cite{mmseg2020} and depth as reinforcement learning based controller's inputs. Note that in grounded experiments, the segmentation is obtained through ground truth.

% \vspace{-0.4cm}
\textbf{Comparison with End-to-end Methods.}
We compare our proposed RSPT with three end-to-end methods (AD-VAT, AD-VAT+, TS), which also utilize RGB-D data as inputs. 
% To note that, we use * to represent the tracker with RGB-D input.
% They regards tracker and target as two learnable agents, and the two agents can make progress together under asymmetric self-play training.
% Especially, the target learns to escape from the visible area of the tracker while the tracker learns to keep up with the target. 
% delete by bx
% In different learning phases, the target people present different escape policies which guide the tracker to cope with various movements of the target. 
% In return, a stronger tracker also forces the target to be optimized. 
It is worthy to notice that the AD-VAT tracker was trained in the \emph{FlexibleRoom} without any obstacles.
Differently, \textbf{AD-VAT+} is an extension of AD-VAT and aims to handle more complex environments with obstacles via a two-stage learning strategy. 
% Thus, a pre-trained AD-VAT tracker is finetuned in the \emph{FlexibleRoom+} environment to learn to cope with obstacles.
% \textbf{AD-VAT+} can be regarded as the state-of-the-art method for active object tracking.
Besides, we construct a variant of \cite{zhong2021distractor}, namely \textbf{TS}, we employ the teacher-student learning framework to train the end-to-end RGB-D tracker. 
% is guided by a teacher tracker, following an imitation learning manner. 
Unlike \cite{zhong2021distractor}, the teacher takes grounded agent-centric grid maps as input for obstacle avoidance instead of relative agent poses.
To ensure a fair comparison, we avoid introducing the distracting player in any of the training environments.
% use the during the competition between the teacher tracker and target, the agents
% It is not predicted to be competitive because it just examines the past trajectory and disregards all other information, such as surrounding structure information.

% \vspace{-0.5cm}
% \textbf{Results Analysis.}

%  For two-stage methods, the VT + Rule, VT + RL, and Seg + RL act only relies on the target pixel position in the current time without considering the surrounding structure and target trajectory trend. The VT + Plan adds path planning to achieve obstacle avoidance. However, It always has a delay. Route planning is carried out according to the current location of the target. When the tracker starts walking according to the navigation, the target has left its previous position. This makes it difficult to maintain a specific distance and angle with the target. 
%  These methods do not achieve good results in obstacle environments. They also cannot be generalized in an unseen environment. 
 %vour method has a higher reward in an environment that has not been seen before, which exceeds the baseline by more than 100 points in \emph{Garden} environments. The episode length of our method is longer compared to baselines. Note that the longest episode length of our method exceeds 100 steps in the \emph{Garden} environment. Our success rate is also higher, the highest in more than 40\% than baselines in the \emph{Garden}.

 The results, as shown in Table~\ref{table-baseline}, demonstrate that our method has a higher reward outperform the best end-to-end baseline by more than 100 points in \emph{Garden}. Our success rate is also significantly higher than baselines in the \emph{Garden} and \emph{SnowForest}.
 The results indicates that it is challenging for end-to-end trackers to extract enough information about the non-current field of view by implicitly encoding raw-pixel observations, leading to suboptimal decision-making. In contrast, using our structure-aware motion representation allows the tracker to develop a better understanding of the surrounding environment and the targets' movements, which is helpful to track in complex environments.
\subsection{Ablation Study}

We design some ablation versions of RSPT,
namely VisT (Visibility of Target), VisT+RS (Reconstruct Structure), VisT+PT (Predict Trajectory), and RSPT (KF). The results are shown in Table~\ref{table-ablation}.
% including the Visibility of Target (VisT), VisT + Reconstruct Structure (RS), VisT + Predict Trajectory (PT), and RS + Kalman Filter (KF). The results are shown in Table~\ref{table-ablation}. 
% to analysis the contribution of each. introduced component in RSPT. The results are shown in Table~\ref{table-ablation}.
% \vspace{-0.03cm}
% ablation studies to analysis the contribution of each introduced component in RSPT. The results are shown in Table~\ref{table-ablation}.
% \vspace{-0.03cm}

% VisT is the most basic component of the structure-aware motion representation and we add different components (RS, PT and KF modules) based on it to construct other three methods.
VisT serves as the foundational component of the state representation, with additional RS, PT, and KF modules added to create three other methods.
Specially, in RS + KF, we replaces PT module in VisT + PT with an off-the-shelf Kalman Filter~\footnote{\url{https://github.com/zziz/kalman-filter}} as our trajectory prediction module. This linear KF filter is used to forecast the future state of a dynamic process using a collection of (low-dimensional) observations. 

% \textbf{RSPT.} 
Compared VisT with the VT + RL results in Table~\ref{table-baseline}, 
VT + RL achieved better performance, indicating that only using the location of the visible target is helpless to improve the generalization of tracker.
% VisT achieved better performance, since the VisT enable to capture the location information in a 3D environment of the target directly. 
% But only relying on it, the tracker is unable to generalize to a complex environment well.
Our results indicate that RS enables the tracker to plan an optimal tracking route around obstacles by leveraging structural information. Meanwhile, PT enables the tracker to adjust its trajectory in anticipation of the target's predicted motion. These improvements highlight the importance of incorporating both structural information and motion prediction in active object tracking.
The performance gap observed between the two ablations (VisT + RS and VisT + PT) and our RSPT (MDN-RNN) demonstrates the effectiveness of the missing components.
% as the tracker can learn to bypass obstacles and keep track based on the map information. When the target goes out of view, the reward experiences a decline in occlusion. Structure information helps the tracker plan to circulate the tracking route from the obstacle, although some cases increase the disappearance time of the target but finally reach a longer episode length.
% When the PT is introduced (VisT + PT), reward significantly improves in all environments compared to the VisT. This improvement results from the tracker adjusting its trajectory in advance using the predicted target motion, thereby keeping the target in the center of its view with an appropriate size.
The performance gap between the two choices of PT module (MDN-RNN vs. KF) underscores the importance of accounting for surrounding obstacles and considering multiple possibilities for the target trajectory in trajectory prediction. 
Our results suggest that the MDN-RNN model's ability to model the target's motion uncertainty and leverage structural information leads to improved trajectory predictions and overall tracking performance compared to the simpler KF module. These findings demonstrate the value of incorporating advanced motion models and structure-aware trajectory prediction.
% Our MDN-RNN based PT module outperforms KF-based module as it considers surrounding obstacles and multiple possibilities of the target trajectory. 
% The combined VisT, RS, and PT modules in our RSPT tracker produce the highest performance. The PT module predicts the target's trend, while the local map helps the tracker plan the route, resulting in a robust tracker in complex environments with structure-aware motion representation.
% According to the result, our RSPT tracker that combines the above VisT, RS, PT modules achieves the highest performance, since the PT module helps to predict the target trend, and the local map helps the tracker to plan the route. Therefore, such a structure-aware motion representation helps produce a robust tracker in complex environments.

%  In addition to the , the RSPT mentioned in other places is RSPT with a target appear state.

% \vspace{-0.1cm}
\subsection{Tracking with Grounded States}
\label{subsec:ground}
% \vspace{-0.1cm}
% To verify the reasonable design of the structure-aware motion representation-based method, we construct two different methods based on different types of grounded state: semantic segmentation masks and depth images.

% \textbf{Object Mask.} It takes the instance-level object segmentation mask and depth image as the inputs. In this experiment, we compare the benefits of structure-aware motion representation and the instance-level scene representation to verify the importance of structure-aware motion knowledge. 

% \textbf{Target Mask.} It takes binary masks of the target and background as well as the depth images as the inputs. Compared to the Object Mask model, this model receives the accurate location of the tracker and achieves a much higher score. However, due to a lack of awareness of environment structure and the target movement, there is still a huge gap between the mask-based methods and RSPT.

% \textbf{RSPT with GT.} We test our RSPT method with ground truth (GT) poses. Note that the poses are from the passive tracker in RSPT. We hope to verify the upper bound of the structure-aware motion representation-based method through this experiment without being affected by the accuracy of the passive tracker. 

% Shown as Table.~\ref{table-gt}, we can see that the structure-aware method performs better in various unseen environments than other methods that use ground truth. This shows that constructing the tracker using our RSPT method can enable the tracker to obtain better robustness in a complex environment.

To identify the bottleneck for AOT, we devised two state representations utilizing different grounded states: semantic segmentation masks and depth images. The first representation, \textbf{Object Mask}, employs instance-level object segmentation masks and depth images as inputs. The second representation, \textbf{Target Mask}, makes use of binary masks of the target and background, as well as depth images as inputs. Additionally, we constructed \textbf{RSPT with GT} by replacing the estimated poses in RSPT (which are based on video tracker and depth image) with ground truth poses. This method serves to verify the upper-bound of the structure-aware motion representation-based approach without being impacted by the accuracy of the video tracker.

% we construct two state representations based on different types of grounded state: semantic segmentation masks and depth images. 1)
% \textbf{Object Mask.} It takes the instance-level object segmentation mask and depth image as the inputs.
% 2) \textbf{Target Mask.} It takes binary masks of the target and background as well as the depth images as the inputs. 
% 3) \textbf{RSPT with GT.} We replace the estimated poses in RSPT (based on the video tracker and depth image) with ground truth (GT) poses to construct this method. In this way, we can verify the upper bound of the structure-aware motion representation-based method without being affected by the accuracy of the video tracker. 

Shown as Table~\ref{table-gt}, the Target Mask model outperforms the Object Mask model due to the binary mask filtering out irrelevant background semantics. This allows the model to focus more on the target and the structure information, leading to improved performance. Despite the Target Mask model's improved performance, it still exhibits significant shortcomings in dealing with various unseen environments due to a lack of awareness of environmental structure and target movement. In contrast, utilizing the proposed RSPT method for constructing the tracker achieved a 100\% success rate in four out of six diverse testing environments, demonstrating that this approach can significantly enhance robustness in complex environments.

% \vspace{-0.1cm}
\subsection{Robustness Analysis}
% \vspace{-0.1cm}
To evaluate the robustness of the trackers, we introduce noise to the input and conduct experiments.
The video tracker is trained on real data, bridging the gap between simulation and reality in RGB processing. However, the depth image captured by a real RGB-D sensor is often noisy, and the mapping method is frequently affected by cumulative errors of the pose, making noise on camera poses a significant concern.
To add noise to the depth image, we use the random-shift/laterally-corruption cooperation from the ICL-NUIM dataset~\cite{handa2014benchmark}. Additionally, we apply Gaussian noise with a mean of $0$ and a variance of $0.5$ meters to the pose to simulate noisy pose.
% However, the remaining gap is mainly due to noisy depth and tracker pose, given that the depth image captured by a real RGB-D sensor is often noisy. Furthermore, the mapping method is frequently affected by the cumulative errors of the pose, making noise on camera poses a significant concern.
% We use the random-shift/laterally-corruption cooperation from ICL-NUIM dataset~\cite{handa2014benchmark} for noisy depth.
% We utilize Gaussian noise with a mean of $0$ and a variance of $0.5$ meters for the noisy pose.
% For the depth image, we employ an off-the-shelf method~\cite{handa2014benchmark, Bohg:etal:2014} to simulate the noise in real depth camera.
% For the camera pose, we simply add Gaussian noise on the grounded pose at each step.

As shown in Table~\ref{table-noise}, our method is not greatly affected by pose errors by constructing tracker-centric surrounding maps. The noise on depth causes more drop in the performance, but our method with noisy input remains achieved higher scores than end-to-end trackers with clear depth.

\begin{figure*}[tb]
  \centering
  \subfigure[In Virtual Environment]{
    \includegraphics[width=0.99\linewidth]{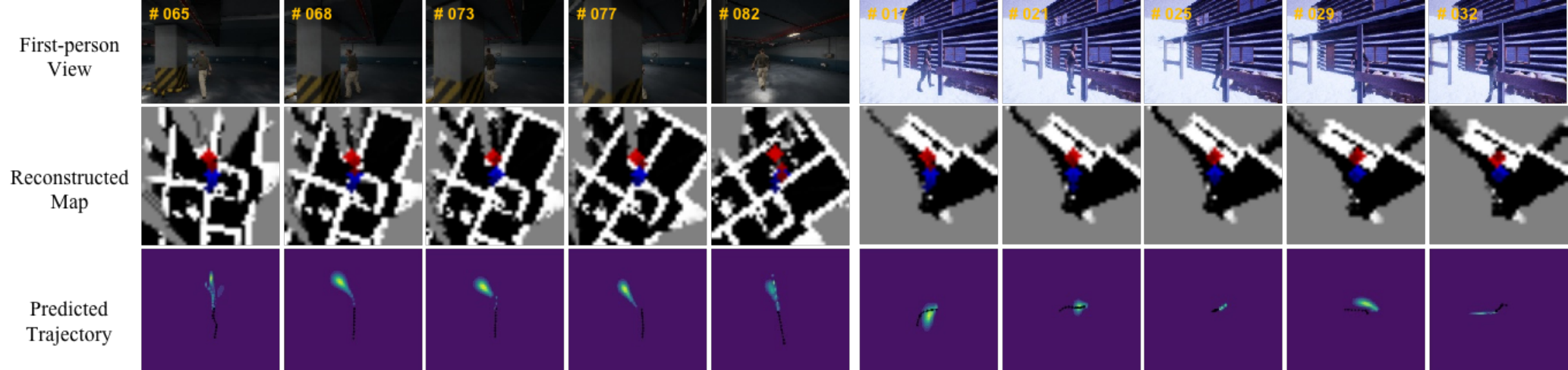}
    }\\%\hspace{4em}
  \subfigure[In Real-world Environment]{
    \includegraphics[width=0.99\linewidth]{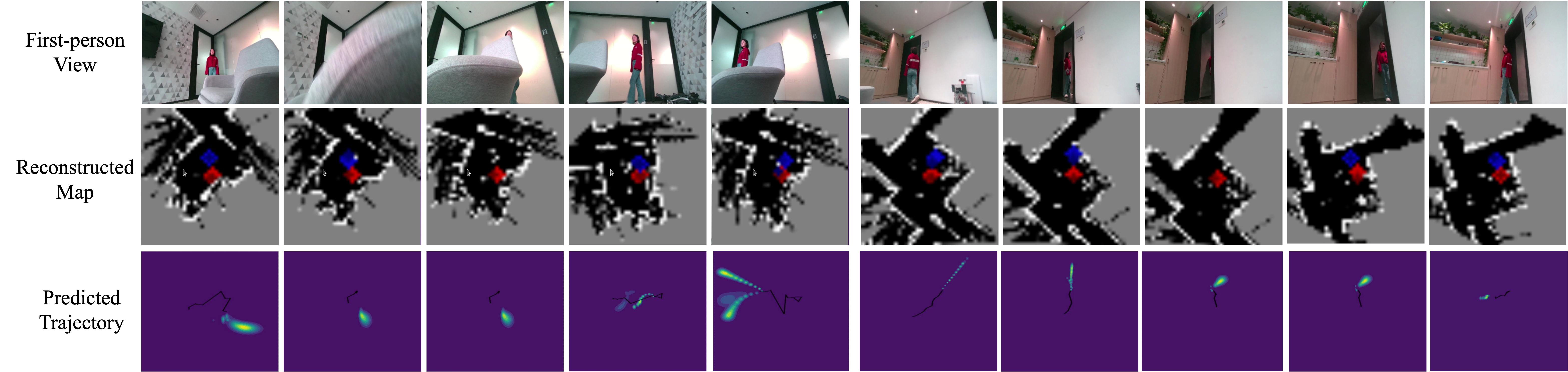}
    }
  \caption{Exemplar sequences of RSPT in different environments.
  The reconstructed map depicts the target as a red dot, the tracker as a blue dot, obstacles as white, free space as black, and the unexplored area as gray. 
  Meanwhile, the predicted trajectory displays the historical trajectory via a black line and the future trajectory distribution via a green area. The latter reflects the probability density function, with brighter colors indicating higher probabilities.}
   \label{fig:cases}
\end{figure*}

% \vspace{-0.1cm}
\subsection{Real-world Deployment}
\label{subsec:real}
% \vspace{-0.1cm}

To verify the practical value of the proposed RSPT tracker, trained in \emph{FlexibleRoom+} with 1m expected distances, we conduct experiments on a physical robot in real-world settings.
The tracker's performance is tested in three situations: Moving Forward, Moving Backward, and Tracking with Visual Occluding. \emph{Moving Forward} involves the target winding around the obstacle or S-type line, testing the tracker's ability in obstacle avoidance and target movement prediction. \emph{Moving Backward} tests the tracker's ability to remember obstacles outside the field of view as the target moves along the reverse direction of the mobile robots into the edge of the barrier. \emph{Tracking with Visual Occluding} tests the tracker's ability to handle visual occlusion as the target hides behind the obstacle.

% The details of the three situations are as followings: 1) Moving Forward: the target winds around the obstacle or S-type line, to test the ability of obstacle avoidance and target movement prediction.
% 2) Moving Backward: the target moves along the reverse direction of the mobile robots into the edge of the barrier to test the memory of obstacles outside the field of view.
% 3) Tracking with Visual Occluding: the target hides behind the obstacle to test the tracker's ability to deal with visual occlusion.

In terms of quantification, we mainly consider the success rate, distance error, and direction error as  ~\cite{luo2019pami}.
We collected $79$ minutes of real-world tracking sequences, a total of $47k$ steps, to evaluate the model. For each frame, we calculate the three indicators using the bounding box, and compute the average value over all frames for each situation. A higher success rate indicates a more robust tracker, while smaller distance error and direction error correspond to higher tracking accuracy.
% For each frame, we use the bounding box to calculate the above three indicators. For each scenario, we calculate the average value over all the frames. The higher the success rate, the more robust the tracker is, and the smaller the distance error and the direction error, the higher tracking accuracy.
% Please refer to the supplementary material for more details.

\begin{table}[tb]
    \centering

    \resizebox{\linewidth}{!}{
    \begin{tabular}{ccccc}
    \hline
        Situation & Method & SR & Distance Error & Direction Error \\ \hline
        \multirow{2}*{MF} & VT + Planner & 0.90 &	0.21±0.05 &	0.13±0.07 \\
        ~ & RSPT & \textbf{0.97} & \textbf{0.14±0.03} &	\textbf{0.08±0.06} \\ \hline
        \multirow{2}*{MB} & VT + Planner & 0.83 &	0.29±0.08 &	0.20±0.09 \\ 
        ~ & RSPT & \textbf{0.91} & \textbf{0.25±0.09} &	\textbf{0.11±0.05} \\ \hline
        \multirow{2}*{VO} & VT + Planner & 0.68 & 0.32±0.10 &	0.37±0.12 \\ 
        ~ & RSPT & \textbf{0.95} & \textbf{0.17±0.06} &	\textbf{0.14±0.08} \\ \hline
    \end{tabular}}
      \caption{We present quantitative results in three real-world situations: Moving Forward (MF), Moving Backward (MB), and Visual Occluding (VO). The three numbers in each cell represent Accumulated Reward (AR), Episode Length (EL), and Success Rate (SR) respectively. We compare our results with the VT + Planner baseline and highlight the best results in bold.}
    \label{table:real}
\end{table}

The quantitative results are shown in Table~\ref{table:real}. \emph{VT + Planner} represents the traditional two-stage method. As the sim2real gap, all the end-to-end methods fail to deploy in our robot.
% which uses Video Tracker to get the target's location, and then use path planning method to navigate to track the target.
Our proposed RSPT tracker outperforms \emph{VT + Planner} in terms of success rate, distance error, and direction error. Specifically, the RSPT tracker achieved a higher success rate in the Moving Forward (MF) situation, which is the most frequently occurring situation in the training data. In addition, even in the Moving Backward (MB) situation with limited space, our approach maintains robust tracking performance. Moreover, our method demonstrates better performance than the traditional method in the Visual Occluding (VO) situation where the 2D tracker is ineffective. This is due to the ability of the RSPT method to utilize the predicted state of the target for control, which enables normal tracking to be maintained. % The target location has been more than 20\% of the traditional method.

% \vspace{-0.2cm}
\subsection{Exemplar Cases}
% \vspace{-0.1cm}
For a better understanding of the workflow of the RSPT tracker, we further provide demo sequences in Figure~\ref{fig:cases}, including simulation and reality scenarios.
% The sequences above are in a simulation. The sequences below are in reality.
% Please watch the demo video for more vivid examples.

%  At every time step, we showed pictures of the first perspective in the first row, the constructed surrounding structure in the second row (red representative target, blue representative tracker, white represents obstacles, black representatives can pass), and the results of trajectory prediction in the third row (black representative historical trajectory, green representing the future trajectory distribution, and the brighter color, the greater the probability). 

% As shown in the figure,the RSPT tracker can make the prediction according to the map and historical position when visual occurs, and continue to track. When the target is hiding in the corner, the tracker can predict the target's intention according to the map information, so the tracker can only wait for the target without moving. Our PT module appears unsatisfactory because it must adapt to various learning targets with widely disparate walking strategies. In future work, we will classify the target and predict its trajectory based on the trajectory pattern.

Figure~\ref{fig:cases} shows that the RSPT tracker can predict the target's position using map and historical trajectory information when visual occlusion occurs, enabling continuous tracking. In cases where the target is hiding in a corner, the tracker leverages map information to predict the target's intention and remains stationary, awaiting the target's emergence. However, the future trajectory proposed by the PT module appears suboptimal in some cases, as it must adapt to diverse learning targets with varying walking strategies. As future work, we plan to classify the target and predict its trajectory based on the pattern of movement.

% \vspace{-0.1cm}
\section{Conclusion}
% \vspace{-0.1cm}
In the field of active object tracking, we assert that accurate prediction of target trajectory and awareness of surrounding obstacles are critical for successful tracking, particularly in environments with clustered obstacles and diverse layouts. To address these challenges, we propose a structure-aware motion representation to develop a generalizable RGB-D tracker. Our approach includes key modules such as target localization, structure reconstruction, structure-aware trajectory prediction, and motion controller. Through experimental results on a range of realistic virtual environments, we demonstrate that our approach exhibits superior generalization performance. Additionally, we confirm the effectiveness of our approach in real-world scenarios by deploying it on a quadruped robot, which shows promising sim-to-real generalization capabilities.
% In the context of active object tracking, we posit that the state of surrounding obstacles and accurate prediction of target trajectory are crucial for successful tracking, particularly in environments with clustered obstacles and diverse layouts. To address these challenges, we propose a structure-aware motion representation to develop a generalizable RGB-D tracker. The key modules of our approach include target localization, structure reconstruction, structure-aware trajectory prediction, and motion controller. Experimental results demonstrate that our approach exhibits superior generalization performance across a range of realistic virtual environments. Furthermore, we validate the effectiveness of our approach in the real-world by deploying it on a quadruped robot, which shows promising sim-to-real generalization capabilities.
% We argue that the state of the surrounding obstacle and the prediction of target trajectory are the most important ingredients for active object tracking, especially in environments with clustered obstacles and diverse layouts. We proposes a structure-aware motion representation to build a generalizable RGB-D tracker. The main modules include target localization, structure reconstruction, structure-aware trajectory prediction and motion controller. Experiments prove that our approach has better generalization in a variety of realistic virtual environments. We also deploy RSPT to the real quadruped robot, showing the good sim2real generalization.

Our work opens up several interesting directions for future research. Firstly, each module of the RSPT tracker can be further developed to address specific challenges. For example, in the structure reconstruction module, it would be important to efficiently build 3D maps for uneven terrain. Additionally, the tracker should be designed to avoid dynamic obstacles such as distractors without colliding with other obstacles. Secondly, for the video tracker, a promising direction is to incorporate the structure representation with image features to improve robustness to distractors and occlusion. Thirdly, for trajectory prediction, we consider adopting the machine theory of mind approach to predict the target's intention~\cite{wang2021tom2c} for more precise trajectory prediction. Furthermore, it would be valuable to explore ways to make the modules mutually beneficial. Lastly, our framework could be extended to more challenging multi-agent settings such as team formation~\cite{jin2022soft}, target coverage control~\cite{xu2020hitmac, pan2022mate}, 3D human pose estimation~\cite{ci2023proactive}. 

% There are also many interesting directions in the future. Each module in RSPT can be further developed. For structure reconstruction, when encountering uneven terrain, the tracker should consider how to efficiently build a 3D map. In addition, when a variety of fast dynamic obstacles, \eg, distractors, appear, the tracker should be sensitive to avoid them without crashing other obstacles. For the video tracker, one feasible direction is incorporating the structure representation with the image feature to enhance the robustness to distractors and occlusion. For trajectory prediction, we consider adopting the machine theory of mind to predict the target's intention~\cite{wang2021tom2c} for more precise trajectory prediction, as each target has its preferences.  It is also worth exploring a way to make the modules mutual benefit. Moreover, it is also a feasible direction to extend the framework for more challenging multi-agent setting, \eg, target coverage control~\cite{xu2020hitmac, pan2022mate}.

% \vspace{-0.2cm}
\section{Acknowledgements}
This work was supported by MOST-2022ZD0114900, China National Post-
doctoral Program for Innovative Talents (Grant No. BX2021008), NSFC-62061136001, and Qualcomm University Research Grant.

\bibliography{aaai23}
\end{document}